\documentclass[times,twocolumn,final]{elsarticle}
\usepackage{medima}
\usepackage{framed,multirow}
\usepackage{amssymb}
\usepackage{latexsym}
\usepackage{url}
\usepackage{xcolor}
\usepackage{hyperref}
\usepackage{amsmath}
\usepackage{amssymb}
\usepackage{comment}
\usepackage{xcolor}
\usepackage{amsmath}
\usepackage{amssymb}
\usepackage{algorithm}
\usepackage{algpseudocode}

\usepackage{array}
\newcommand{\bfx}{\mathbf{x}}
\newcommand{\bfz}{\mathbf{z}}
\newcommand{\bfv}{\mathbf{v}}
\newcommand{\bfV}{\mathbf{V}}
\newcommand{\argminA}{\arg\min}

\usepackage{soul}

\newcommand{\rj}[1]{{\textcolor{orange}{Rohit: #1}}}

\definecolor{newcolor}{rgb}{.8,.349,.1}

\journal{Medical Image Analysis}

\begin{document}

\verso{Yifan Wu \textit{et~al.}}

\begin{frontmatter}

\title{Neural Ordinary Differential Equation based Sequential Image Registration for Dynamic Characterization} 
%\tnoteref{tnote1}}%\tnotetext[tnote1]{This is an example for title footnote coding.}

\author[1]{Yifan \snm{Wu}\corref{cor1}}
\cortext[cor1]{Corresponding author: Y. Wu (yfwu@seas.upenn.edu)}
%\fntext[fn1]{This is author footnote for second author.}
\author[1]{Mengjin \snm{Dong}}
\author[1]{Rohit \snm{Jena}}
\author[2]{Chen \snm{Qin}}
\author[1]{James C. \snm{Gee}}
%\author[]{Alzheimer’s Disease Neuroimaging Initiative\corref{cor2}}
%\cortext[cor2]{Data used in preparation of this article were obtained from the Alzheimer’s Disease Neuroimaging Initiative (ADNI) database (\url{adni.loni.usc.edu}). As such, the investigators within the ADNI contributed to the design and implementation of ADNI and/or provided data but did not participate in analysis or writing of this report. A complete listing of ADNI investigators can be found at: \url{http://adni.loni.usc.edu/wp-content/uploads/how_to_apply/ADNI_Acknowledgement_List.pdf}}

%\author[1]{James \snm{Gee}\fnref{fn1}}

\address[1]{Penn Image Computing and Science Laboratory, University of Pennsylvania, Philadelphia, PA 19104, USA}
\address[2]{Department of Electrical and Electronic Engineering and I-X, Imperial College London, UK}
%\address[2]{Affiliation 2, Address, City and Postal Code, Country}

\received{N/A}
\finalform{N/A}
\accepted{N/A}
\availableonline{N/A}
\communicated{N/A}

\begin{abstract}
Deformable image registration (DIR) is crucial in medical image analysis, enabling the exploration of biological dynamics such as organ motions and longitudinal changes in imaging. Leveraging Neural Ordinary Differential Equations (ODE) for registration, this extension work discusses how this framework can aid in the characterization of sequential biological processes. Utilizing the Neural ODE's ability to model state derivatives with neural networks, our Neural Ordinary Differential Equation Optimization-based (NODEO) framework considers voxels as particles within a dynamic system, defining deformation fields through the integration of neural differential equations. This method learns dynamics directly from data, bypassing the need for physical priors, making it exceptionally suitable for medical scenarios where such priors are unavailable or inapplicable. Consequently, the framework can discern underlying dynamics and use sequence data to regularize the transformation trajectory. We evaluated our framework on two clinical datasets: one for cardiac motion tracking and another for longitudinal brain MRI analysis. Demonstrating its efficacy in both 2D and 3D imaging scenarios, our framework offers flexibility and model agnosticism, capable of managing image sequences and facilitating label propagation throughout these sequences. This study provides a comprehensive understanding of how the Neural ODE-based framework uniquely benefits the image registration challenge.

\end{abstract}

\begin{keyword}
\KWD Deformable Image Registration\sep Sequential Registration\sep Registration for Dynamic Analysis\sep Flow-based registration\sep Neural Ordinary Differential Equation
\end{keyword}

\end{frontmatter}

\section{Introduction}
Deformable image registration (DIR) aims to establish spatial correspondence between images and serves as a crucial tool in medical image analysis for characterizing the dynamics of biological systems. These dynamics include organ motions, changes in longitudinal imaging, and more. While most existing algorithms perform registration on a pair of discrete observations \cite{sotiras2013deformable}, namely the moving and fixed images \cite{sotiras2013deformable}, the dynamics are essentially continuous over time. In this work, we explore how Neural Ordinary Differential Equation (ODE)-based registration frameworks can enhance the characterization of dynamics.

Neural Ordinary Differential Equations (ODE) were proposed by Chen \emph{et al.} \cite{chen2018neural}. Drawing inspiration from ResNet \cite{he2016deep}, which employs residual networks to model differences between hidden states, Neural ODE assumes that the step between consecutive states is infinitesimally small. It then uses neural networks to parameterize the derivative of states, thus theoretically modeling continuous dynamics. Both the final and the intermediate states, when available, are utilized to supervise the optimization of network parameters. 

In our previous work, we introduced the Neural Ordinary Differential Equation Optimization-based (NODEO) framework for image registration \cite{Wu_2022_CVPR}, tailoring the generalized flow field approach specifically for the medical registration domain. This framework conceptualizes each voxel as a moving particle and views the collective set of voxels in a 3D image as a dynamic system. The trajectories of these particles, determined through the integration of neural differential equations, define the desired deformation field. These equations model the relationship between the velocities and positions of all particles across various time points, with the initial condition being that the spatial transformation is the identity. Unlike conventional approaches that leverage neural networks for feature extraction from vast datasets \cite{VoxelMorph_v1, VoxelMorph_v4, mok2020fast, dalca2019learning, yang2017quicksilver}, our NODEO framework leverages the expressive capabilities of neural networks to parameterize ordinary differential equations. 

In this work, we discuss how this registration framework can describe the dynamics of biological systems, and/or use such dynamics to constrain spatial transformation results. Image registration, inherently an ill-posed problem, typically relies on deriving deformations from image intensity information. The choice of the registration model and objective function implies an assumption on the nature of deformation and constraints on the solution space with such prior. Most existing approaches impose uniform regularization such as smoothness and diffeomorphism \cite{mok2020fast, shen2019networks}. To explicitly model the dynamics, flow-based models such as elastic body models \cite{gee1993elastically} and viscous fluid flow models \cite{d2003viscous}, describe the physical process of the deformation using a dynamics-defined differential equation. Yet, these physical priors may not always be readily available or applicable in real-world medical scenarios, including cases like infant development \cite{wang2019developmental}, disease progression \cite{adler2018characterizing}, and movements due to cardiac or respiratory functions. 

Theoretically, Neural Ordinary Differential Equations (ODEs) offer a novel approach by obviating the need for physical priors. Instead, they employ neural networks to formulate the differential equations, thereby learning to identify the dynamics directly from the data itself. In scenarios where sequential images exsist to model temporal effects, most existing methods rely on pair-wise solutions, treating each pair of images as an independent and identically distributed (i.i.d) observation. Unlike these methods, our framework not only generates the final spatial transformation between the moving and fixed images but also delineates the trajectory along the sequence. This approach allows intermediate images to act as state constraints for the trajectory, thus offering prior information about the underlying dynamics.

Experimentally, our framework can naturally apply to a pair or a sequence with an arbitrary number of images without adding model complexity. It provides a feature to propagate initial labels throughout the sequence, significantly reducing the effort required for labeling in situations where only a few segmentation labels might be available. Moreover, the framework is flexible, allowing for easy configurations in optimization objectives and network selection to suit specific applications and data types. Additionally, it accommodates adaptive time step sizes based on the physical time intervals, enhancing its applicability and precision across various temporal sequences.

In this work, we present a generic registration framework and demonstrate its properties and advantages using relevant data. Our empirical study encompasses two clinical datasets: the ACDC cardiac sequence dataset \cite{bernard2018deep}, consisting of 2D images for motion analysis, and the ADNI dataset for MRI biomarkers of AD \cite{petersen2010alzheimer}, comprising 3D images for longitudinal analysis. These datasets facilitate a comprehensive evaluation of our framework's capability to handle both 2D and 3D images across different medical imaging scenarios, demonstrating its utility dynamic characterization.

\section{Related Works}
\subsection{Sequential Registration}
One of the earliest algorithms capable of sequential registration is the LDDMM framework~\cite{beg2005computing} which parameterizes diffeomorphic transformations as the solution of a viscous flow PDE with a time-dependent velocity field, and solves the Euler-Langrage equations for image matching.
The velocity field is stored as an explicit 4D tensor representing velocities as a function of discretized spatial grid coordinates and time.
An Euler integration scheme is applied to compute the diffeomorphism, and without any other regularization, the velocity field is optimized to obey the geodesic constraint.
However, this method is prohibitively expensive for pairwise registration due to the storage complexity of the 4D velocity field.
% ~\cite{sandkuhler2019recurrent} present a recurrent network for deformable image registration, 
Sequential registration methods are also utilzed to analyze the dynamics of biological systems.
~\cite{qin2018joint} proposed a sequential registration method to analyze the biomechanics of the heart. They used a biomechanical model to simulate the heart's motion and used sequential registration to estimate the model parameters.
The prior on sequential registration essentially learns the dynamics of the myocardium which is essential for motion tracking.
However, most sequantial algorithms in practice are often modelled as a sequence of pairwise registrations. 
~\cite{ko1997multiresolution} use a feature-based sum of absolute differences method combined with a coarse-to-fine strategy to register coronary arterial images. 
Sequential image registration is also used to extract mycardial motion and deformation from tagged MRI (t-MRI) imaging, where the sequantial registration paradigm is chosen to compensate for tag fading over time~\cite{morais2013cardiac}.
However, actual registration is still done in a pairwise manner.

%  In the medical imaging field, sequential registration is used to analyze the progression of diseases, such as Alzheimer's disease \cite{wang2011alzheimer}, multiple sclerosis \cite{ceccarelli2010quantitative}, and infant brain development \cite{gousias2012automatic}.
%  In these applications, the registration of multiple images from the same subject is essential to monitor the changes over time.

\subsection{Neural ODE on Registration}
Since many approaches model the diffeomorphism as solution of a velocity-flow PDE, a viable approach to registration is to learn the parameterization of the PDE (i.e. the velocity field). 
This velocity field may or may not depend on time. Existing literature aims to embed priors and constraints on the velocity field to ensure desirable properties of the diffeomorphism.
\cite{Han_2023_WACV} proposes a neural network (MLP) to parameterize a stationary velocity field that parameterizes the diffeomorphism.
This choice of representation exploits implicit regularization provided by the network weights. 
A cascade-style registration is performed by choosing a FCN to perform an initial registration which is refined by the MLP.
\cite{xu2021multi} propose a multi-scale ODE network that performs registration by solving ODEs at different resolutions. This is done to improve convergence speed and avoid local minima associated with a single-scale optimization.
Neural ODEs have applications beyond uniform image grids.
Deep Implicit Functions are used to represent 3D geometry with continous representations, and a neural diffeomorphic flow is used to maintain the topology of the geometry represented by the implicit function~\cite{Sun_2022_CVPR}.
This method can represent a quasi time-varying velocity field by learning K velocity fields instead, in similar spirit to LDDMM.
Similar to our work,~\cite{sun2022medical} propose a neural field represented by a MLP, which is used to predict either a deformation field or a velocity field that is integrated to obtain the diffeomorphism.
The neural ODE stems from the neural representation of the velocity field used in ~\cite{sun2022medical}.
~\cite{van2023echocardiography} propose a 2D latent UNet for encoding the image and segmentation into a transformation from the first image in the sequence to the image at time $t, t \in [0, 1]$. The sequential nature of the the echocardiogram provides intermediate images as `checkpoints' for the registration.
Therefore, a time dependent velocity parameterization is used, and neural ODE is used to integrate over this sequence of transforms.
% A similar approach is also used by ~\cite{wolterink2022implicit} to learn a deformation vector field (DVF) for pairwise registration. 
A related application of Neural ODEs is to perform data-driven modeling of nonlinear, anisotropic materials for finite element analysis~\cite{tac2022data}.
This enables accurate modeling of tissue dynamics in biomechanical simulations, for example, myocardium segmentation sequences in ~\cite{qin2020biomechanics}.
Neural ODEs have also been used to perform compositional representation of point clouds in latent deformation spaces.
~\cite{Jiang_2021_CVPR} propose a model with a compositional encoder which encodes identity, pose, and motion. Two point cloud pairs from a sequence are randomly picked from th 4D sequence and fed into the encoder. A neural ODE is then used to update the pose encoding to update the final pose of the point cloud from the initial pose.
In this case, the deformation is implicit via the pose encoder but the neural ODE is used to learn dynamics rather than predict them.

\subsection{Registration for Dynamic Analysis}

\subsubsection{Registration for Motion Analysis}
Sequential image registration has been explored for many clinical applications. 
For lung images, statistical motion models can be learned from data and generating warp fields using registration algorithms.
~\cite{ehrhardt2010statistical} propose a method to generate a mean motion model of the lung based on thoracic 4D computed tomography (CT) data of different patients.
Registration of spatiotemporal 4D CT scans of the lung can be used for functional investigations ~\cite{ford2003respiration,guerrero2006dynamic}.
~\cite{reinhardt2008registration} describe a registration-based technique to estimate a local lung expansion model from respiratory-gated CT images of the thorax.
The Jacobian of the displacement field is used to represent local tissue expansion.
% This is compared to Xe-CT based measures of lung ventilation.
In the Xe-CT data, they compute the mean and standard deviation of the Xe-CT sV parameter capturing measures of lung ventilation. 
Registration accuracy is then evaluated by comparing volume expansion with a functional measure of lung ventilation.
~\cite{mcclelland2006continuous} use sequential non-rigid registration along with a respiratory signal to calculate the position of the respiratory cycle for each free breathing CT scan.
A B-spline function is then used to map the position in the respiratory cycle with the displacement of the control points from a reference breath-hold scan.
~\cite{ruhaak2017estimation} propose a registration field based on similarity evaluation of normalized gradient fields of a sparse set of keypoints over a predefined search space to initialize and refine a dense registration.
This leads to an algorithm that achieves excellent alignment of lung vessels, airways and fissures leading to accurate lung motion model estimation.
~\cite{low2005novel} propose a motion model as linear functions of tidal volume and airflow.
The displacement at a location $\overrightarrow{r}$ is described as the sume of two independent displacement vector components: tidal volume and airflow. 
In addition to the CT scans, spirometry-measured tidal volume was simultaneously acquired, and a fitted to a fifth-order polynomial to avoid unstable derivative estimates.
A linear relationship between the tidal volume and airflow is used to learn the motion model.
% ~\cite{sundaram2005towards} estimate the kinematics 
Early works on heart motion analysis ~\cite{tagare1999shape} used a shape-based nonrigid correspondence algorithm to match plane curves.
An objective function for comparing local shape is used and the algorithm is validated using return error of tracked points, defined as the distance between the tracked point over one heart beat period.
Other works~\cite{chandrashekara2003construction,mcleish2002study,bustin20203d,krebs2021learning,zhu2021test} have constructed various statistical models for motion analysis using sequential registration.
More recently, ~\cite{qin2018joint,qin2020biomechanics,qin2023generative} use finite element methods to generate phantom motion sequences and use this prior to regularize the registration of cardiac images.

\subsubsection{Registration for Longitudinal Analysis}
Longitudinal image analysis plays a crucial role in monitoring the progression of specific regions by utilizing multiple scans per subject \cite{pegueroles2017longitudinal, dong2021deepatrophy}. This approach finds wide applications in fields such as infant brain development analysis, tracking changes in multiple sclerosis \cite{vrenken2013recommendations}, and observing brain atrophy in normal aging or Alzheimer's disease (AD) patients \cite{cash2015assessing, xie2020longitudinal}. Within Alzheimer's disease studies, longitudinal analysis of brain atrophy serves as a vital biomarker on the structural level, enabling us to comprehend disease progression, identify early signs of Alzheimer's, and investigate the link between brain atrophy and cognitive decline \cite{jack2018nia, adler2018characterizing}.

Traditionally, longitudinal analysis methods for AD progression on T1-weighted structural MRI fall into two main categories: deformation-based morphometry (DBM) \cite{hua20083d, das2012measuring} and boundary shift integral (BSI) \cite{freeborough1997boundary, prados2015measuring}. In DBM, an image registration technique generates a voxel-wise deformation field or Jacobian determinant from an image pair of the same subject, which is then masked by the hippocampus area – one of the earliest shrinking areas in the brain – to calculate an overall atrophy rate. On the other hand, BSI quantifies the shrinkage area as the intensity difference or shift of hippocampus boundaries between an image pair. Recently, deep learning algorithms have been developed, employing self-supervised learning techniques to learn changes between images in hidden layers \cite{ouyang2022self}, or utilizing time-related information to extract progressive changes from the image \cite{dong2021deepatrophy, dong2023regional}.

Despite these advancements, the mentioned algorithms can only analyze one image pair at a time. To overcome this limitation and enable the analysis of multiple images for a subject, a linear mixed-effect model is typically employed to derive a summarized atrophy measurement \cite{xie2020longitudinal, hua20083d}. In our work, NODEO introduces sequential registration, allowing for an arbitrary number of images as input, reducing the impact of random noise, and obtaining more reliable and consistent hippocampus volumes at each time point. This novel approach promises to enhance the accuracy and precision of longitudinal brain atrophy analysis.

\section{Method}

%%% Rohit: Copied to notes
% \todo{ %
% 1. change all pairwise settings to from 1 to T,  where T can be 2 or N. 
% 2. Mention adaptable t, but we need careful about continuous/discrete discussion.
% 3. Loss function: 
% similarity: MSE, NCC
% Regularity: Jacobian: no folding and cross 
% Application Prior on J: volume preservation or decreasing
% NEW: 
% 3.1 DIR formulation
% 3.2 Neural ODE equations
% 3.3 DIE in dynamical system view
% 3.4 loss function 
% 3.5 network choice 
% complexity analysis (comparison between sequential and pairwise) (Move to experiment)
% }

\subsection{Formulation of Sequential deformable Image Registration}
Deformable Image Registration (DIR) is generally formulated as an optimization problem.
Consider an unparameterized N-dimensional image as a function $I: \Omega \rightarrow \mathbb{R}^d$.
The image takes the location of the i$^{th}$ voxel/pixel, denoted as $\bfx_i \in \Omega \subset \mathbb{R}^N$ and converts it into an intensity value $I(\bfx_i)$.
% Given two images $I$ and $J$ $: \Omega \rightarrow \mathbb{R}^d$, the deformation $\phi: \Omega \rightarrow \Omega$ transforms the image $J$ by $J \circ \phi$.
For brevity, we consider 3D scalar images in this scenario, i.e., $N=3, d=1$, although the formulation is applicable for any positive $N$ and $d$.
A deformation field, on the other hand, is given by $\phi: \Omega \rightarrow \Omega$ which transforms points in the image domain.
DIR is the problem of finding a deformation $\phi^*$ such that the transformation $J \circ \phi^*$ is close to $I$.
Since $J$ is transformed, we denote $J$ as the \textit{moving image} and $I$ as the fixed image.
Therefore, DIR can be formulated as the following optimization problem:
\begin{equation}
   \phi^* = \argminA_{\phi} \mathcal{S}(J\circ\phi, I) + \mathcal{R}(\phi)
\end{equation}
where $\mathcal{S}$ is a dissimilarity function between images $J$ and $I$, and $\mathcal{R}$ is a prior on $\phi$.
In many applications of DIR, a \textit{desirable} property of $\phi$  is to be sufficiently smooth and diffeomorphic in $\Omega$.
The latter condition requires that the topology of the moving function is preserved under transformation.
In other words, the transformation $\phi$ should not create folds in $\Omega$.
This can be formulated by a regularizer $\mathcal{R}$ on $\phi$.
\\

In sequential registration, the notion of deformation is extended to be time-dependent, $\phi: \Omega\times[0, \infty) \rightarrow \Omega$.
Moreover, given a sequence of $T$ time points $0 < t_1 < t_2 \ldots < t_T $, we have $T$ fixed images $F_{i}$ corresponding to time $t_i$ and a moving image $M$ corresponding to $t =0$.
With slight abuse of notation, let $\phi_t = \phi(\mathbf{\cdot}, t)$.
The sequential registration problem can be now formulated as 
finding the optimal deformation $\phi^*$ by minimizing:
\begin{equation}
    \phi^* = \argminA_{\phi} \sum_{i=1}^{T} \mathcal{S}\left(M\circ \phi_{t_i}, F_i \right)  + \mathcal{R}(\phi)
\end{equation}

Similar to previous works \cite{Wu_2022_CVPR}, many choices of dissimilarity functions are possible. The most commonly used ones are the mean squared error (MSE), normalized cross correlation (NCC), global and local mutual information (MI).
% For regularizing the deformation field, 
Diffeomorphism in registration is typically encouraged by penalizing the determinant of the Jacobian, since a negative Jacobian indicates folding and crossing in the deformation field.  
Moreover, application-specific priors can be built into the regularization term.
For example, volume preservation can be encouraged by forcing the determinant of the Jacobian to be close to 1, shrinkage in volume can be encouraged by a hinge loss $H(\mathcal{J}) = \max{\left(|\mathcal{J}| - 1, 0\right)}$, penalizing Jacobians that only locally expand the volume, hyperelastic terms \cite{qin2020biomechanics}, etc.
% Deformation fields can also be implicitly regularized by parameterizing them \todo{CITE}~\cite{deepimageprior}, and by formulating them as the solution of a differential equation \todo{cite}~\cite{lddmm}.

Extending the DIR formulation from a pair of images to a sequence brings an additional challenge - parameterizing the deformation field.
A straightforward way to formulate deformations in sequential DIR is to consider independent deformations $\phi_i$ for registering images $M$ and $F_i$.
Deformation fields can also be parameterized by formulating them as the solution of a differential equation ~\cite{beg2005computing}.
In this paper, we take the latter approach, by formulating the time derivative of the deformation field as a parameterized, learnable velocity field, and optimizing its parameters instead.
% This allows us to capture underlying dynamics of the sequence of images, which 
Unlike independent pairwise optimization, this formulation allows us to discover and capture the underlying dynamics that is shared across the image sequence, by accumulating registration errors from all time points.
Moreover, the number of parameters doesn't scale with number of time points in the image sequence, which subsequentially improves optimization time.

\subsection{Neural Ordinary Differential Equations}
Taking inspiration from the resemblance between residual networks and dynamical systems, Chen et al.~\cite{chen2018neural} first introduced neural ordinary differential equations (NODEs) to approximate infinite depth neural networks. It aims to learn the function $f$ parameterized by $\theta$ by defining a loss function of the following form
\begin{equation}
\begin{aligned}
    \frac{dz}{dt} = f_{\theta} (z(t), t),
    \label{eqn: NODE1}
\end{aligned}
\end{equation}

\begin{equation}
\begin{aligned}
    \mathcal{L}(\bfz(t_1)) = \mathcal{L}\left(\bfz_0 + \int^{t_1}_{t_0} f_{\theta}(\bfz(t), t)dt\right).
    \label{eqn: NODE2}
\end{aligned}
\end{equation}

From a systems perspective, NODEs are continuous-time models that represent vector fields as neural networks. It has since been adapted as a universal framework for modeling high-dimensional spatiotemporally chaotic systems utilizing convolutional layers~\cite{Jiahao2021KnowledgebasedLO}, demonstrating its ability to capture highly complex behaviors in space and time. 
% Hence, we find it a suitable candidate for our registration task. 
NODEs ~\cite{chen2018neural} have been shown to perform better than recurrent networks both in interpolation and extrapolation, when the dynamics is consistent for the time sequence data.
Since sequential and pairwise DIR are ill-posed problems alike, one way of constraining the solution space is to discover underlying shared dynamics using a parameterized velocity field.
This can produce deformations that are more `consistent' than their independent pairwise registration counterparts, which do not share learned dynamics for different time points.  

Since NODEs often require the numerical solver to take many steps to realize the flows, they are memory-inefficient if all gradients along the integration steps need to be stored using traditional backpropagation. Hence many recent works~\cite{chen2018neural, zhuang2020adaptive} on NODEs have therefore focused on reducing the memory requirements for gradient propagation.
Notably, the adjoint sensitivity method (ASM) has enabled constant memory gradient propagation for optimizing NODEs, and we adopt ASM in our work as well. For a brief description of ASM, one can refer to the supplementary. Proofs for its gradient convergence can be found in~\cite{chen2018neural, Jiahao2021KnowledgebasedLO}. 
ASM enables our framework to interpolate between $t=0$ and $t=T$ for an arbitrary number of steps with a constant memory cost.
This is particularly helpful when a temporally smooth diffeomorphic flow is required, as the numerical solver can increase its number of steps to improve the smoothness of $\phi_t$ with respect to $t$.

\subsection{DIR in Dynamical System View}
Our work borrows intuition from dynamical systems and treats the trajectory of the entire deformation field as the solution to a first-order non-autonomous ordinary differential equation given by
\begin{equation}
\begin{aligned}
    \frac{d\phi}{dt}(\bfx, t) = \mathcal{K} \bfv_{\theta} (\phi(\bfx, t), t),\\ 
    s.t.\ \phi_0 = \text{{Id}},
    \label{eqn: psi diff}
\end{aligned}
\end{equation}
where $\bfv_{\theta}(\cdot, \cdot)$, parameterized by $\theta$, is the velocity vector field describing the dynamics of the deformation field.
% voxel cloud, %$\Pi$ is the state space, and 
$\phi_0$ is the initial condition at $t=0$. 
We employ 3D Gaussian kernels ($\Omega \subseteq \mathbb{R}^3$), denoted by $\mathcal{K}$, as a filtering operator to enforce spatial smoothness in $\bfv_\theta$. 
% Intuitively $\mathcal{K}$ is to ensure that the velocities of voxels are similar to those of their neighbors.
Intuitively, $\mathcal{K}$ is to ensure that the velocity field is not too jagged.
%\rj{Following sentence is not very useful, we all know what gaussian kernel does.}
Increasing the kernel size amounts to smoothing over larger voxel space, and therefore will improve the smoothness of the resulting flow; increasing the variance of the kernel amounts to encouraging more individual movements and therefore reduces the smoothness of the resulting flow.

Since sequential image registration may have time-dependent dynamics (atrophy of hippocampus increases with time \cite{dong2021deepatrophy}), the time derivative of $\phi$ (i.e., the velocity field) explicitly depends on $t$.
% 
% The term \textit{non-autonomous}, or equivalently \textit{time-variant}, \textit{non-stationary} means that the time derivative of $q$ explicitly depends on $t$~\cite{strogatz2018nonlinear}. In other words, the velocity field attached to the Eulerian frame changes with time. 
% 
The trajectory of $\phi$ is generated by integrating the ODE in Eq. \eqref{eqn: psi diff} with the initial condition $\phi_0$. 
% Assuming that the voxel cloud evolves from $t=0$ to $t=s$, the resulting voxel cloud at $t=s$ denotes the transformation $\psi(q_0)$ given by
The evolution of $\phi$ from $t=0$ to $t=T$ is given as 

\begin{equation}
    % \psi(q_0) = q(s) = q_0 + \int^{s}_{0} \mathcal{K} \bfv_{\theta}(q(t), t) dt.
    \phi(\bfx, t) = \phi(\bfx, 0) + \int_{0}^{T} \mathcal{K} \bfv_{\theta}(\phi(\bfx, t), t) dt
    \label{eqn: int}
\end{equation}

The optimization problem therefore becomes:
\begin{equation}
\begin{aligned}
\small
    \theta^* = \argminA_{\theta \in \Theta} \mathcal{L}_{sim}\left(I, J(\phi_0 + \int^{T}_{0} \mathcal{K} \bfv_{\theta}\left(\phi(\bfx, t), t\right) dt)\right)\\
    + \mathcal{R}(\phi, \bfv_{\theta})  + \mathcal{B}(\phi),
    \label{eqn: our optim}
\end{aligned}
\end{equation}
where $\Theta$ is the space of all possible parameters. The different components in the loss function include the similarity metric $\mathcal{L}_{sim}$, the regularizers $\mathcal{R}$, and the boundary conditions $\mathcal{B}$. 

The primary difference with other methods is that since image sequences may have time dependent dynamics, an explicit dependence on $t$ is required, and further reduction or simplication of the velocity field is not possible like %\todo{CITE}
~\cite{beg2005computing}.

\subsection{Choice of network representation}
In the aforementioned framework, $\bfv_\theta(\bfx, t)$ is a function of a spatial location $\bfx $ and $t$.
This parameterization limits our representation capacity to usage of pointwise MLPs only.
On the contrary, network architectures like ConvNets take a grid as input, but do not fit into the framework.
To overcome this, we define the function $\bfV_\theta$ to be a function $\bfV_\theta: 2^\Omega\times[0,\infty) \rightarrow 2^\Omega$ whose 
domain and range are subsets of $\Omega$ instead.
Now consider the set $\mathcal{P} = \{\phi(\bfx, t) | \bfx \in \Omega\}$. 
From the definition of $\phi$, we have $\mathcal{P} \subset \Omega$.
We define
\begin{equation}
    \bfV_\theta(\mathcal{P}, t) = \{ \bfv_\theta(p, t) | p \in \mathcal{P} \}
    \label{eq:overload v}
\end{equation}
Intuitively, Eq.~\ref{eq:overload v} operates on the entire set of transformed point at any given time, and outputs the corresponding velocities at all these points.
This allows us to use arbitrary architectures such as ConvNets, which takes the entire grid of $\phi(\bfx, t)$ as input, and outputs a corresponding grid of $\bfv_\theta$.
We overload $\bfv_\theta$ with $\bfV_\theta$ and use this representation in Eq. ~\ref{eqn: our optim}.
Note that the resulting substitution does not guarantee a diffeomorphic transform in $\phi$ anymore.
%\rj{Yifan, please add the explanation about why its not diffeomorphic anymore, from the previous paper}.
To mitigate this problem, we add regularizers to encourage diffeomorphism, which is described in Section~\ref{subseq:loss}.

In this work, we use two types of networks 
\begin{itemize}
    \item a UNet-like convolutional neural network, which takes a grid of transformed points as input. %\rj{Yifan please add more details}
The ConvNet operates as follows:
At time $t$, let $G_t$ be a grid of size $H\times W\times D$ where $G_t[i, j, k] = \phi(\bfx_{ijk}, t)$.
The ConvNet takes the concatenation of $G_t$ with a positional encoding of $t$ in the channel dimension as input.
It computes an output grid $H_t$ of the same size as $G_t$, where $H_t[i, j, k] = \bfv_\theta(\bfx_{ijk}, t)$.
    \item A per-voxel MLP, which takes a point $\phi(\bfx, t)$ and $t$ as input, and outputs $\bfv(\phi(\bfx, t), t)$.
    % This parameterization can guarantee diffeomorp
    Under certain conditions of boundedness and step size of the numerical integration \cite{miller2006geodesic}, this formulation can be guaranteed to be diffeomorphic.
    However, using feedforward architectures with standard non-linearities has shown to be unable to capture high frequency features \cite{sitzmann2020implicit}.
    
    %To alleviate these issues, we use a SIREN network with 64 hidden layers.The spatial coordinates $\phi(\bfx, t)$ and time $t$ are separately passed through a SIREN layer to get high frequency features.These features are concatenated and passed through a SIREN network to output the final velocity. More details are provided in the supplementary. 
\end{itemize}

The choice of architecture induces an implicit prior on the velocity field \cite{ulyanov2018deep}. However, an in-depth investigation of this implicit bias and its properties is beyond the purview of our current study. Our framework is agnostic to the network architecture, allowing for flexible adaptation to suit various applications and datasets.

\subsection{Loss functions}
\label{subseq:loss}

A variety of functions can be used for $\mathcal{L}_{sim}$s, $\mathcal{R}$ and $\mathcal{B}$ depending on the modality and application. 
In this work, the dissimilarity measure $\mathcal{L}_{sim}(I, J) = 1 - NCC(I, J)$ is used, where $NCC$ is the normalized cross correlation given by:

\begin{equation}
\begin{array}{l}
{NCC}(I, J)=\\
\frac{1}{N} \sum_{\bfx \in q(s)} \frac{\sum_{\bfx_{i}\in W}(I(\bfx_{i})-\bar{I}(\bfx))(J(\bfx_{i})-\bar{J}(\bfx))}{\sqrt{\sum_{\bfx_{i}\in W}(I(\bfx_{i})-\bar{I}(\bfx))^{2} \sum_{\bfx_{i}\in W}(J(\bfx_{i})-\bar{J}(\bfx))^{2}}},
\end{array}
\label{eqn: NCC}
\end{equation}
where $\bar{I}(\bfx)$ and $\bar{J}(\bfx)$ are the local mean of a size $w^3$ window $W$ with $\bfx$ being at its center position, and $\bfx_{i}$ is an element within this window. In this work we set $w$ to be 21.

Other dissimilarity measures like mean squared error, or mutual information can be used as well.

The regularization term consists of three terms: 
\begin{equation}
\label{eqn: R}
    \mathcal{R}(\psi, \bfv_\theta) = \lambda_1\mathcal{L}_{Jdet} + \lambda_2\mathcal{L}_{mag} + \lambda_3\mathcal{L}_{smt}.
\end{equation}
The first term, $\mathcal{L}_{Jdet}$, penalizes negative Jacobian determinants in the transformation and is given by 

\begin{equation}
% \mathcal{L}_{Jdet}=\frac{1}{N} \sum_{\bfx \in \phi_t}  \|\sigma(-(|\mathcal{D}_{\phi_t}(\bfx)|+\epsilon))\|_{2}^{2},
\mathcal{L}_{Jdet}=\frac{1}{N} \sum_{\bfx \in \Omega}  \|\sigma(-(|\mathcal{D}{\phi(\bfx, t)}|+\epsilon))\|_{2}^{2},
\label{eqn: Jdet}
\end{equation}
where $\mathcal{D}{\phi}$ is the Jacobian matrix at $\bfx$ under the transformation $\phi$. Here $\sigma(\cdot)=max(0, \cdot)$ is the ReLU activation function, which is used to select only negative Jacobian determinants.
If there are no %holes or 
folds in the transformation, its Jacobian $\mathcal{D}{\phi}(\bfx, t)$ should have a positive determinant.
Lastly, we add a small number $\epsilon$ to the Jacobian determinants as an overcorrection. 
Instead of using $L1$ regularization as in \cite{mok2020fast,VoxelMorph_v4}, we use $L2$ norm here. Regularization with $L1$ introduces sparsity, reducing the number of folds, while the $L2$ norm can minimize the overall magnitude of folds, thereby avoiding outliers. To adapt to a specific task, one can combine the two.
% In our framework, $\mathcal{L}_{Jdet}$ is a critical component since it ensures that the flow is diffeomorphic in $\Omega$.
In this work, the Jacobian matrix is implemented using the finite difference approximation.

\subsection{Comparison of Pairwise and Sequential Registration}
%In the context of sequential image registration problems, the methodology adopted for temporal registration can significantly impact both the computational efficiency and the fidelity of the registration process. Traditional pairwise registration methods involve initializing and optimizing an independent transformation for each pair of consecutive images (i.e., between time $t$ abd $t+1$). This method results in a rapid escalation of computational resources as the sequence extends, posing a limitation in terms of scalability.

%Conversely, the sequential registration framework presents a model that encapsulates both efficiency and temporal coherence. By leveraging a consistent set of parameters $N$, which parameterized by a neural network, the complexity of the model does not escalate with the addition of frames. This attribute is facilitated through the formulation of a differential equation, which sequentially adapts the transformation parameters over time. Such a strategy not only simplifies the parameter space but also ensures model scalability regardless of sequence length.

In the context of sequential image registration problems, the methodology adopted for temporal registration can significantly impact both the computational efficiency and the fidelity of the registration process. Unlike traditional pairwise methods that require a new set of transformations for each consecutive image pair, which quickly increasing computational demand and treat each pair independently. Our sequential registration formulation employs a constant parameter set $N$, parameterized by a neural network. This network represents a differential relation between velocity field and deformation field at time $t$, managing model complexity efficiently. This approach maintains temporal coherence and model scalability, significantly streamlining the parameter space without being scaled by sequence length (Table~\ref{table:comparison}).

A significant advantage of sequential registration is its utility in delineating a transformation pathway between a start and endpoint, particularly when intermediate frames are present. This methodology is capable of producing not just the final deformation but also a series of intermediate deformations. These intermediate deformations offer a granular view of the transformation process over time, serving as a comprehensive trajectory of the registration path.

Furthermore, the inclusion of intermediate frames inherently imposes constraints on the deformation trajectory, which act to regularize the transformation. This regularization is instrumental in maintaining the physical or logical consistency of the sequence, thereby enhancing the overall accuracy of the registration process.

Importantly, the sequential registration approach does not compromise on precision. Empirical evaluations demonstrate that its accuracy is on par with direct registration methods between the initial and final frames, despite its reduced model complexity. This retention of accuracy, alongside the benefits of model efficiency and the insights gained from intermediate deformations, positions sequential registration as a more fit method for handling temporal sequences in image registration tasks.

\begin{comment}
\begin{algorithm}
\caption{Conceptual registration on sequential images}
\begin{algorithmic}
\State Let $I$ be the fixed image and $J$ the moving image. Let $J^0 = J$ and $\phi^0 = \text{Id}.$
\For{$i = 0, \dots, M$}
    \State Compute optical flow $v^i$ between images $I$ and $J^i$
    \State Let $\phi'^i = \text{Id} + \epsilon v^i$, i.e., $\phi'^i(x) = x + \epsilon v^i(x)$
    \State Let $\phi^{i+1} = \phi^i \circ \phi'^i$, i.e., $\phi^{i+1}(x) = \phi^i(x + \epsilon v^i(x))$
    \State Let $J^{i+1} = J^0 \circ \phi^{i+1}$
    \State Terminate if the difference between $J^{i+1}$ and $J^i$ is below a threshold.
\EndFor
\end{algorithmic}
\end{algorithm}
\end{comment}

\section{Experiments}
\subsection{Registration for Cardiac Motion Tracking}
\subsubsection{Registration Task Introduction}
The motion and deformation of the myocardium are valuable indicators of cardiac function. 
Image registration is the predominant technique utilized for motion tracking analysis, enabling the determination of anatomical correspondence between frames. Motion tracking and segmentation are interdependent and commonly solved jointly \cite{qin2018joint}, with segmentation propagation derivable from registration results. 

\begin{table}[t]
\centering
\begin{tabular}{|c|m{6cm}|}
\hline
\textbf{Type} & \centering\arraybackslash \textbf{Description} \\
\hline
& \centering\arraybackslash $\phi_t = \phi_{12} \phi_{23} \phi_{34} \dots \phi_{(t-1)t}$ \\
 & \centering\arraybackslash Each $\phi_{(t-1)t}$ is parameterized by \\
 Pairwise        &  \centering\arraybackslash $\theta_{(t-1)t}$ and is optimized independently. \\
         & \centering\arraybackslash Total number of parameters: N $\times$ T \\
\hline
 & \centering\arraybackslash $\phi_t = \phi_{(t-1)} + f_\theta(\phi_{(t-1)}, t) \cdot \delta t$ \\
  Sequential         & \centering\arraybackslash The function $f_\theta$ represents a differential equation. \\
           & \centering\arraybackslash Total number of parameters: N \\
\hline
\end{tabular}
\caption{Comparison of pairwise and sequential methods.}
\label{table:comparison}
\end{table}

Registration algorithms in this task are commonly evaluated on two factors: 
the accuracy of the registration and the regularity of the resulting deformation fields \cite{qin2018joint, qin2020biomechanics, qin2023generative}. The accuracy is assessed using the Dice Similarity Coefficient (Dice) and Mean Contour Distance (MCD). Dice evaluates the overlap between the warped source image segmentation and target ground truth segmentation, while MCD measures the average distance between the contours of the warped source segmentation and the target segmentation. A higher Dice score and a lower MCD indicate better accuracy. The regularity of the deformation fields in this task is defined based on the prior knowledge that myocardium volume is preserved during motion. This is assessed using the Jacobian Determinant ($\| J|-1|$),
which represents the mean absolute difference between the Jacobian determinant ($\operatorname{det}\left(J_{\Phi}(x)\right)$) and $1$, denoted as $\| J|-1|$.  A lower $\| J|-1|$ indicates better preservation of volume.

\begin{figure*}[t]	
\small
	\begin{center}
		\includegraphics[width=\linewidth]{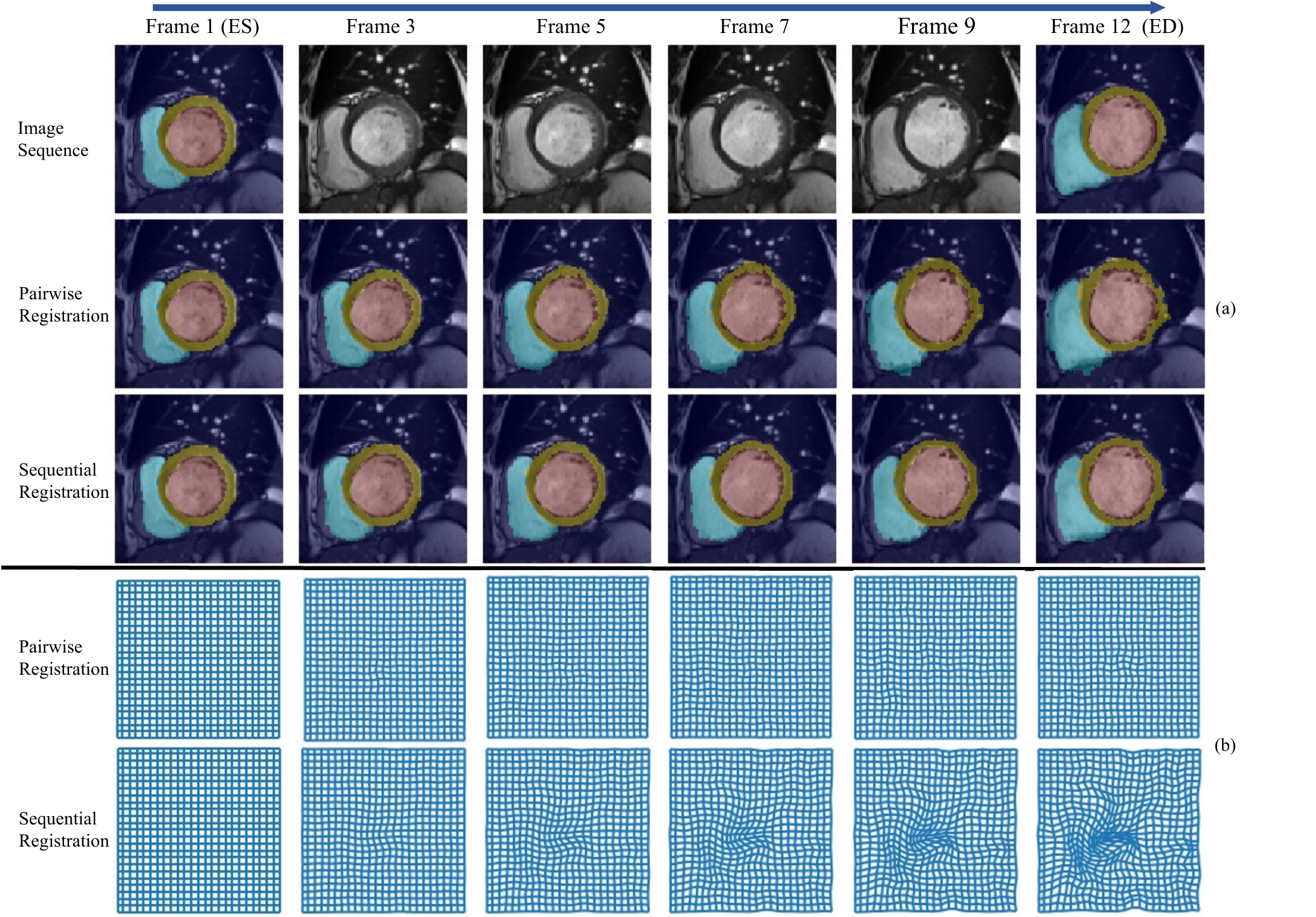}
	\end{center}
	\caption{The qualitative results of label propagation via registration along sequence. The example shown here is patient 001 in ACDC dataset, which has 12 frames from ES to ED. The ground-truth segmentation labels are only available for frames ED and ES. In (a), we show original image sequence with ground truth segmentations of right ventricle (RV), left ventricle (LV) cavities, and the myocardium in the first row. The second row and the third row demonstrate qualitative comparison between pairwise registration, which conducts registration between $t$ and $t+1$ for $t=1, 2, ..., N-1$, where $N=12$, and sequential registration where the one-time registration processes the whole sequence. Here the frames are subsampled for visualization. In (b), we show the corresponding deformation field for both pairwise and sequantial registration.} %\todo{show the composition of deformation here for pairwise registration.}}
    \label{fig: label_ropagation}
\end{figure*}

\subsubsection{Dataset and Data Processing}
We use the ACDC dataset in this application study.
The ACDC dataset \cite{bernard2018deep} is a widely utilized collection for cardiac image analysis including cardiac motion analysis, segmentation, function analysis, and diagnosis. It comprises 100 cardiac magnetic resonance imaging exams, each containing cine-MRI sequences ranging in length from 12 to 35 frames. Expert-annotated segmentation labels for the left ventricle (LV), right ventricle (RV), and myocardium (MYO) are available on two frames in each exam sequence, the end-diastolic (ED) and end-systolic (ES) frames. These exams were collected from a single center and represent a diverse patient population with various cardiac pathologies.

The ACDC dataset has a range of spatial resolutions ranging from $1.37$ to $1.68$ millimeters. In line with related state-of-the-art work \cite{qin2023generative}, we resample each image to a resolution of $1.8 \times 1.8$ millimeters and perform experiments on three representative image slices, as recommended by \cite{taylor2015myocardial} and \cite{schuster2015cardiovascular}. These slices are selected at 25\%, 50\%, and 75\% of the LV length, representing the basal, mid-ventricle, and apical planes, respectively. By evaluating the tracking performance on these slices, we aim to gain a comprehensive understanding of the algorithm's behavior across different planes of the cardiac volume.

\subsubsection{Properties and Advantages of Our Framework}
\label{Sec: exp_properties}

\noindent\textbf{Model Efficiency on Example of Label Propagation.} In this example, the entire sequence is available, but only the end-systolic (ES) and end-diastolic (ED) frames have segmentation masks available. By conduct sequential registration, we can obtain the segmentation propagation, as shown in Figure~\ref{fig: label_ropagation}. 

As concluded in Table~\ref{table:comparison}, our model complexity doesn't increase increase with the number of frames, in contrast with pairwise registration. Moreover, we compare the registration result quantitatively in Table~\ref{table:label_propagation}, which indicates that sequential registration here is not only more efficient, but also with better performance.

\begin{table}[t]
\caption{Comparison of pairwise and sequential registration results on the basal slice of the first 10 patients in the ACDC dataset using our framework. MCD (mm) and Dice scores are calculated on the ED and warped ES slices, with lower MCD and higher Dice indicating better accuracy.}
	\centering
	\resizebox{\linewidth}{!}{
		\begin{tabular}{c|cccc}
			\hline\hline
			Method &MCD (LV)$\downarrow$& MCD (RV)$\downarrow$ &Dice (LV)$\uparrow$& Dice (RV)$\uparrow$ \\ 
            \hline
			Pairwise& $2.445(0.865)$ & $3.075(0.865)$ & $0.753(0.073)$ & $0.842(0.070)$ \\          
			Sequence& $1.556(0.356)$ & $1.869(0.640)$ & $0.838(0.032)$ & $0.909(0.045)$ \\
			
			\hline
		\end{tabular}
    }
	\label{table:label_propagation}
\end{table}
\begin{table*}[!ht]
\centering
\scalebox{0.8}{
\small
\begin{tabular}{cc|ccc|ccc|ccc}
\hline
\hline
 \multirow{2}{*}{Dataset} &  \multirow{2}{*}{Method} & \multicolumn{3}{c}{ Apical } & \multicolumn{3}{c}{ Mid-ventricle } & \multicolumn{3}{c}{ Basal }\\
\cline { 3 - 11 } & & MCD$\downarrow$ & Dice$\uparrow$ & $\| J|-1|$$\downarrow$ & MCD$\downarrow$ & Dice$\uparrow$ & $\| J|-1|$$\downarrow$ & MCD$\downarrow$ & Dice$\uparrow$ & $\| J|-1|$$\downarrow$\\
\hline \multirow{7}{*}{ ACDC } 
& FFD-VP 
& $2.784(1.648)$ & $0.684(0.143)$ & $0.148(0.075)$ 
& $2.442(1.352)$ & $0.756(0.088)$ & $0.147(0.072)$ 
& $2.925(1.328)$ & $0.731(0.120)$ & $0.135(0.052)$
\\
& dDemons 
& $2.437(1.453)$ & $0.707(0.128)$ & $\mathbf{0.132}(0.043)$
& $1.993(1.112)$ & $0.788(0.075)$ & $0.141(0.0needs48)$
& $2.295(1.174)$ & $0.778(0.102)$ & $0.132(0.031)$
\\
& Motion-Net 
& $2.853(1.301)$ & $0.656(0.142)$ & $0.171(0.060)$ 
& $2.799(1.007)$ & $0.745(0.105)$ & $0.167(0.056)$ 
& $2.814(1.236)$ & $0.751(0.123)$ & $0.167(0.049)$ 
\\
& BINN 
& $2.450(1.253)$ & $0.707(0.147)$ & $0.158(0.081)$ 
& $2.210(0.918)$ & $0.783(0.097)$ & $0.148(0.056)$ 
& $2.229(0.860)$ & $0.789(0.091)$ & $0.161(0.063)$ 
\\
& SOTA 
& $2.101(1.312)$ & $\mathbf{0.731}(0.137)$ & $\underline{0.134}(0.091)$ 
& $1.646(0.971)$ & $\mathbf{0.818}(0.055)$ & $\mathbf{0.134}(0.067)$ 
& $\mathbf{1.660}(0.671)$ & $\mathbf{0.829}(0.067)$ & $\mathbf{0.138}(0.060)$
\\
& Ours-Pair 
& $\underline{1.851}(1.133)$ & $\underline{0.715}(0.131)$ & $0.143(0.112)$ 
& $\mathbf{1.542}(0.551)$ & $\underline{0.805}(0.073)$ & $0.162(0.103)$ 
& $\underline{1.707}(0.845)$ &  $\underline{0.802}(0.084)$  & $0.151(0.086)$
\\
& Ours-Sequence
& $\mathbf{1.780}(0.832)$ & $0.702(0.119)$ & $0.148(0.114)$
& $\underline{1.618}(0.545)$ & $0.788(0.072)$ & $\underline{0.138}(0.078)$ 
& $\mathbf{1.660}(0.547)$ &  $0.793(0.078)$  & $\underline{0.149}(0.084)$
\\
\hline
\end{tabular}
}
\caption{\label{Table:benchmark} Quantitative comparisons of registration performance among FFD-VP (\cite{rohlfing2001intensity}), dDemons (\cite{vercauteren2007non}),
Motion-Net (\cite{qin2018joint}), BINN (\cite{qin2020biomechanics}), Generative-BINN (\cite{qin2023generative}), and our proposed method on pair (ES and ED) and sequence (from ES to ED). Results are measured between ED and warped ES frames. The performance of each method is evaluated using three measures: MCD (mean curvature distance), Dice coefficient, and $\| J|-1|$. The results are reported as the mean (standard deviation). The methods with the best and second-best performances are indicated in \textbf{Bold} and \underline{underline}, respectively. }
\end{table*}

\noindent\textbf{Sequence Aware Trajectory.} In this study, we compare the performance of our proposed method for registration of cardiac MR exams to standard benchmarks. The benchmark methods include classical optimization-based approaches such as free-form deformation with volumetric preservation (FFD-VP) \cite{rohlfing2001intensity} and diffeomorphic Demons (dDemons) \cite{vercauteren2007non}, as well as three state-of-the-art data-driven learning-based approaches: Motion-Net \cite{qin2018joint}, the biomechanics-informed neural network (BINN) \cite{qin2020biomechanics}, and the advanced Generative BINN model (Generative-BINN) \cite{qin2023generative}.

\begin{figure*}[t]	
\small
	\begin{center}
		\includegraphics[width=0.95\linewidth]{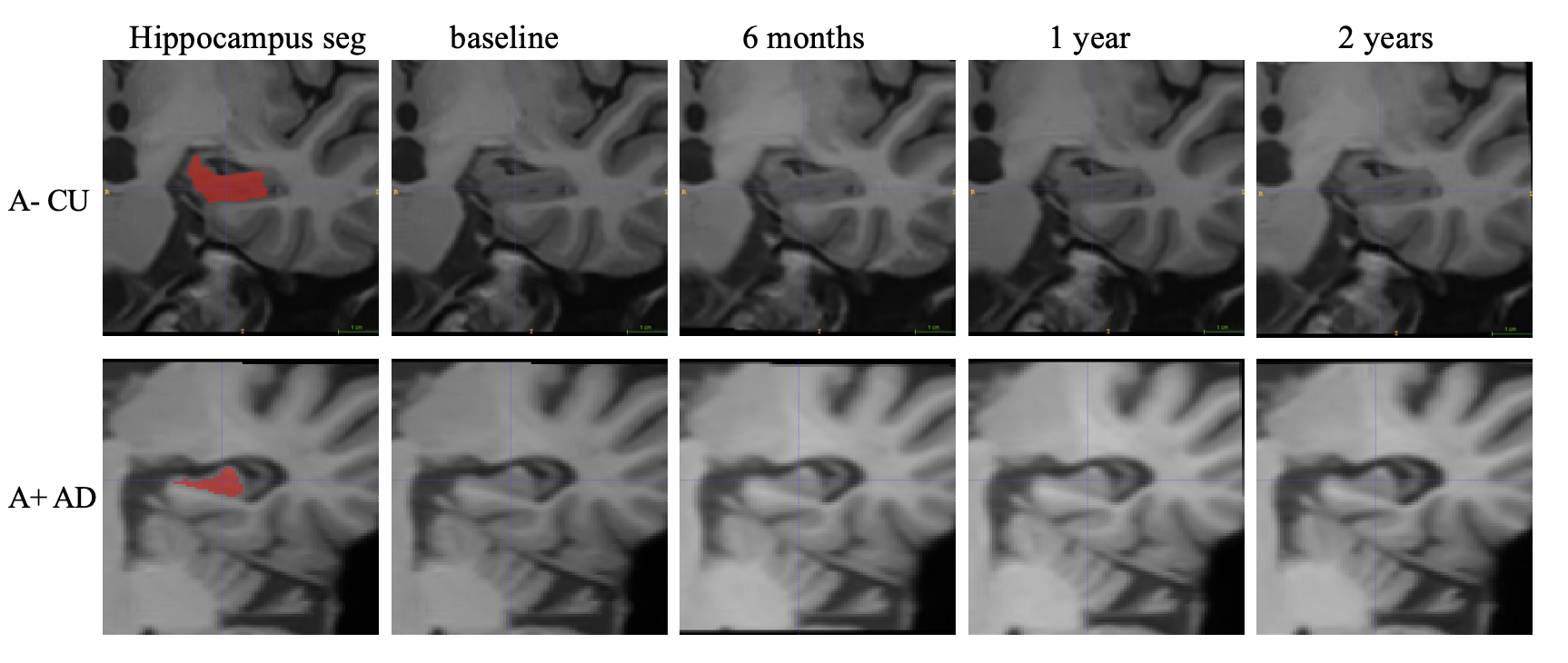}
	\end{center}
	\caption{Comparison of hippocampal images between two disease stages: Amyloid-negative cognitively normal (A- CU) participants and Amyloid-positive Alzheimer's disease (A+ AD) participants. The A+ AD group typically exhibits smaller hippocampi and accelerated hippocampal volume loss compared to the A- CU group.}
    \label{fig: ADNI_example}
\end{figure*}

Image registration is a problem that is considered ill-posed, meaning that it does not have a unique solution. In order to obtain a physiologically plausible solution, assumptions must be made about the deformation fields. To constrain the solution space, FFD-VP and dDemons add uniform and generic regularization terms, such as smoothness, diffeomorphism, and incompressibility, into the optimization objective. To incorporate application-specific priors, Motion-Net employs an approximation of Huber loss to penalize displacement gradients, while BINN and Generative-BINN explicitly incorporate biomechanical knowledge about cardiac motion. BINN achieves this through an explicit regularization term, while Generative-BINN encodes these priors in an embedding space using simulation data and performs test-time optimization on real data. Additionally, Generative-BINN uses test-time optimization, which significantly improves the overall performance and results in the best performance on the benchmark among the compared methods.

The benchmark methods perform pair-wise image registration, where the ED (end-diastolic) frames serve as the fixed images and the ES (end-systolic) frames serve as the moving images. In contrast, our framework is capable of performing registration on the entire sequence of images in a single step. To quantitatively compare the performance of our framework with the benchmark methods, we evaluate both pairwise registration (i.e., registering the ES frames to the ED frames) and sequential registration (i.e., registering the ES frames to the ED frames along the entire sequence) using our framework, as presented in Table \ref{Table:benchmark}. When performing registration on the entire sequence of images, the intermediate images serve as state constraints for the trajectory of the deformation field. Our framework uses the data itself as prior information about the underlying dynamics and constrains the final deformation accordingly. 

Even though our method is a generic image registration framework that is not specifically designed for cardiac imaging or trained on a dataset for this task, both our pairwise and sequential registration methods still achieve results that are comparable to state-of-the-art methods. For all apical, mid-ventricle, and basal slices, our method achieves better accuracy in terms of the mean contour distance and comparable volume preservation in terms of $||J|-1|$. We demonstrate that our framework is able to register the sequential data without loss of performance, and can not only produce the final deformation field, but also the entire trajectory along the sequence.
%\subsubsection{Temporal Interpolation and Extrapolation}

\subsection{Longitudinal Registration for hippocampus atrophy estimation}

\subsubsection{Registration Task}

\begin{figure}[t]	
\small
	\begin{center}
		\includegraphics[width=\linewidth]{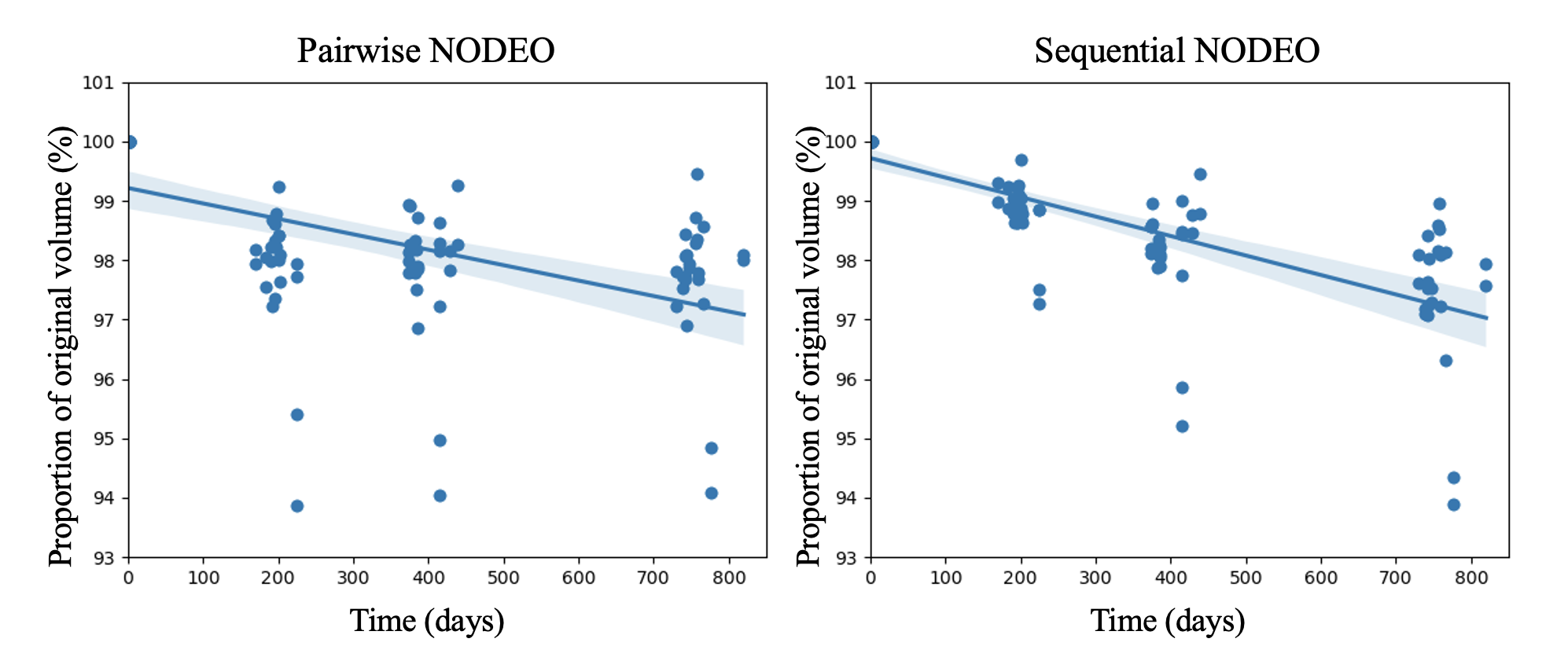}
	\end{center}
	\caption{Comparison of volumetric changes in the hippocampus between pairwise and sequential NODEO analyses in the Amyloid positive Alzheimer's disease (severe AD) group on ADNI. Volumes from each scan are normalized by the baseline volume, yielding proportions relative to the baseline volume expressed as percentages.}
    \label{fig: ADNI_seq_vs_pair}
\end{figure}

Alzheimer's disease is a neurodegenerative disease characterized by the loss of neurons that leads to brain shrinkage and cognitive decline in later stages \cite{dubois2007research, jack2018nia}. Compared to cognitively unimpaired individuals who also experience neuronal loss in the brain, AD patients often exhibit accelerated neuronal loss (see ~\ref{fig: ADNI_seq_vs_pair}). This loss is initially observed in the medial temporal lobe, particularly the hippocampus, and gradually spreads throughout the brain \cite{adler2018characterizing}. The longitudinal analysis of hippocampal atrophy emerges as a powerful tool to detect the earliest signs of atrophy and monitor disease progression. Compared to treating each MRI scan separately and obtaining independent volumes for each MTL area, studies have shown that using a deformable image registration to obtain deformation fields, and then derive the amount of change from the deformation field would significantly reduce variance and lead to more accurate atrophy measurement. \cite{das2012measuring}

Due to the small size of the hippocampus and relatively low resolution of the MRI imaging technique, a direct measurement of brain volumes from MRI scans cannot be considered as the ground-truth measurement (of course, volumes of the hippocampus and MTL cannot be measured \textit{in vivo} either). So it would not be possible to directly compare DICE between image segmentation and "ground-truth" segmentation. This leads to alternative methods to evaluate the performance of longitudinal analysis techniques \cite{cash2015assessing}. In our study, atrophy rates are calculated separately for the Amyloid negative cognitively unimpaired (A- CU) group and the Amyloid-positive Alzheimer's disease (A+ AD) group, and a difference between these two groups is compared to show the sensitivity of NODEO to different rates of changes in the hippocampus. In addition, assuming that hippocampal volume shrinks linearly over time \cite{cash2015assessing, dong2021deepatrophy}, we fit a linear regression line for each diagnosis group using all available scan times. The correlation coefficient and variance of each linear fit are calculated. A high correlation coefficient and low variance demonstrate a high level of consistency in tracking disease progression.

\begin{figure*}[t]	
\small
	\begin{center}
		\includegraphics[width=\linewidth]{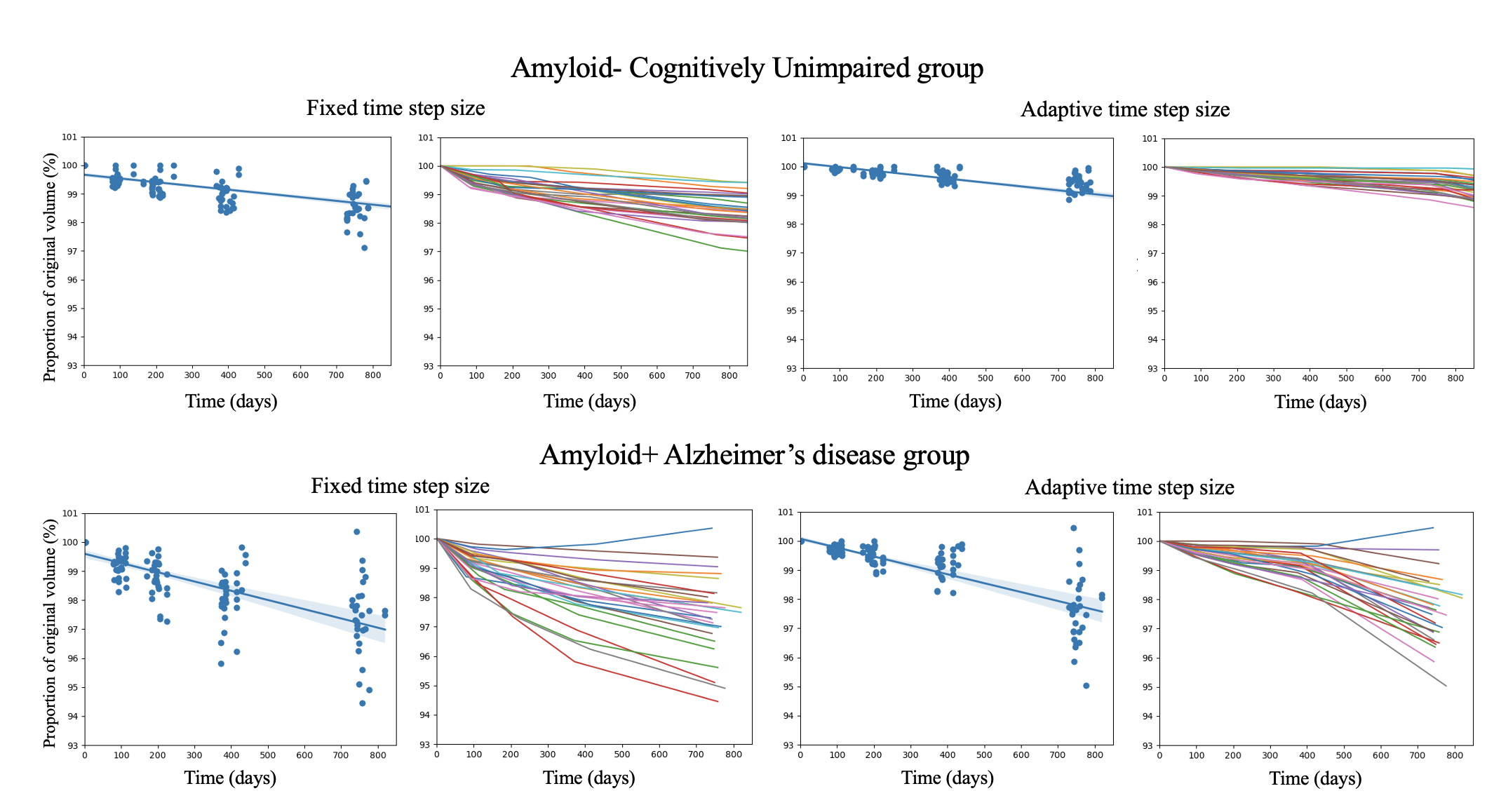}
	\end{center}
	\caption{Comparison of progression trajectories among subjects using fixed and adaptive time-step sizes NODEO, as well as between Amyloid-negative cognitively unimpaired and Amyloid-positive Alzheimer's disease groups. Spaghetti plots illustrating individual participant progressions are depicted with colorful lines for each hemisphere of the hippocampus.}
    \label{fig: ADNI_adaptive_time_steps}
\end{figure*}

\subsubsection{ADNI Dataset and Processing}

Data used in this study were obtained from the Alzheimer’s Disease Neuroimaging Initiative (ADNI, \url{adni.loni.usc.edu}). The ADNI was launched in 2003 as a public-private partnership, led by Principal Investigator Michael W. Weiner, MD. The primary goal of ADNI has been to test whether serial magnetic resonance imaging (MRI), positron emission tomography (PET), other biological markers, and clinical and neuropsychological assessment can be combined to measure the progression of mild cognitive impairment and early Alzheimer’s disease. For up-to-date information, see \url{www.adni-info.org}.

T1-weighted MRI data from 28 participants were selected from ADNI2 and ADNI-GO datasets. Out of these participants, 14 were from A- CU group, and the remaining 14 were from A+ AD group. Each participant underwent a total of 5 MRI scans over 2 years (initial scan, 3 months, 6 months, 1 year, and 2 years).

For the baseline image (the first scan for each subject), the left and right hippocampus were segmented using the ASHS-T1 method \cite{xie2019automated}. Subsequently, the hippocampus area in the image was trimmed by 10 voxels in each direction and cropped to focus on the region of interest. All follow-up images (scans conducted at later times for each subject) were rigidly registered to the baseline image using a greedy registration method \cite{yushkevich2016ic} and cropped accordingly to ensure consistency in the region of interest. All cropped images were upsampled to a resolution of 0.5x0.6x1mm using bilinear interpolation.

\subsection{Implementation Details}

In this longitudinal study, we compared pairwise and sequential registrations conducted using NODEO. In pairwise registration, all follow-up images (i.e., images obtained after the initial scan) were registered to the baseline scan using NODEO. The volumes of the baseline images are focused on early onset areas of AD such as the hippocampus and MTL \cite{xie2019automated}, and that of the follow-up images were obtained by warping and integrating the baseline segmentation to corresponding follow-up images. For sequential registration, we implemented and compared both fixed time-step and adaptable time-step NODEOs. In fixed time-step NODEO, the time interval ($\delta t$) between scans remained constant, regardless of differences in interscan intervals. In adaptable time-step NODEO, however, interscan intervals were factored into the algorithm, and the number of time steps for NODEO for each scan pair was adjusted in proportion to the real interscan interval between two images.

\subsubsection{Properties and Advantages of Our Framework}

\noindent\textbf{Sequence Aware Trajectory.} Figure~\ref{fig: ADNI_seq_vs_pair} illustrates the volumetric changes in the hippocampus for both pairwise and sequential NODEO analyses in the A+ AD group. All follow-up image volumes are normalized by the baseline volume, and the resulting proportion of volumes at each scan interval relative to the original volume is presented. In pairwise NODEO, a significant decrease in volume is evident between the baseline image and the 6-month scans, with minimal volumetric differences observed at subsequent follow-up intervals (6 months, 1 year, and 2 years). This discrepancy may be attributed to the presence of measurement noise, which could overshadow hippocampal volume changes in pairwise registration. Sequential registration effectively mitigates this issue. In the NODEO model, deformation fields are optimized by a regularization term in addition to the data similarity term, potentially reducing the impact of random noise in intermediate scans and resulting in a more linear progression compared to pairwise NODEO. The correlation coefficients for the linear fits in pairwise and sequential registrations are 0.28 and 0.56, respectively, indicating the superior progression estimation capability of sequential NODEO compared to pairwise registration. \\

\noindent\textbf{Adaptable Time Step Size.} In Figure~\ref{fig: ADNI_adaptive_time_steps}, we present a comparison of scatter plots, linear fits, and spaghetti plots for each individual (treating left and right sides as separate observations) using NODEO sequential registration between the A- CU and A+ AD groups, while also optimizing NODEO settings for fixed and adaptive time step sizes. The average atrophy for the A- CU group is 1.34\% and 1.31\% for fixed and adaptive time-step sizes, respectively, while for the A+ AD group, it is 3.05\% and 3.18\% over a two-year period. These findings align with the progression rates reported in the literature for these two diagnostic groups, albeit slightly on the lower side \cite{cash2015assessing}. The observed differences in atrophy between the two diagnostic groups underscore the validity of NODEO measurements as indicative of volumetric changes rather than random factors in the images. Furthermore, adaptive time step size does not significantly affect overall volumetric measurements over two-year intervals, but it does reduce the variability in volumetric measurements in intermediate scans. Taken together, these findings demonstrate that NODEO is a suitable algorithm for longitudinal analysis and biomarker tracking.

\section{Conclusion}
We extend the Neural Ordinary Differential Equation-Based Optimization Framework for Deformable Image Registration (NODEO-DIR) to sequential data. We discuss how this framework identifies the underlying dynamics and uses the sequential data to regularize the transformation trajectory. Our experiments on the registration of cardiac data for motion tracking, and brain data for longitudinal studies, demonstrate the framework's properties and advantages. Specifically, our framework offers efficiency gains; the neural network serves as a compressed representation of the dynamical changes along the sequence, and the model complexity does not increase with the number of frames. Moreover, our framework demonstrates an advantage in that the solution trajectories are sequence-aware, where the deformation is better constrained by the underlying dynamics.
\section{Acknowledgements}
This work was supported in part by grants R01-HL133889, R01-EB031722, and RF1-MH124605 from the NIH.

\bibliographystyle{model2-names.bst}\biboptions{authoryear}
\bibliography{refs}
\end{document}